\documentclass[preprint,12pt]{elsarticle}

\usepackage{amssymb}

\journal{Digital Signal Processing}

\usepackage{xcolor}
\usepackage{hyperref}
\usepackage[caption=false]{subfig}

\usepackage{xspace}
\newcommand{\ie}{i.e.,\xspace}
\newcommand{\eg}{e.g.,\xspace}

\begin{document}

\begin{frontmatter}



\title{An Overview of Computational Approaches for Interpretation Analysis}




\author{Philipp Blandfort}
\address{TUK and DFKI, Kaiserslautern, Germany}

\author{J{\"o}rn Hees}
\address{DFKI, Kaiserslautern, Germany}

\author{Desmond U. Patton}
\address{Columbia University, NYC, USA}

\begin{abstract}

It is said that beauty is in the eye of the beholder.
But how exactly can we characterize such discrepancies in interpretation?
For example, are there any specific features of an image that make person A
regard an image as beautiful while person B finds the same image displeasing?
Such questions ultimately aim at explaining our individual ways of interpretation,
an intention that has been of fundamental importance to the social sciences from the beginning.
More recently, advances in computer science brought up two related questions:
First, can computational tools be adopted for analyzing ways of interpretation?
Second, what if the ``beholder'' is a computer model,
i.e., how can we explain a computer model's point of view?
Numerous efforts have been made regarding both of these points,
while many existing approaches focus on particular aspects and are still rather disconnected.

With this paper, in order to connect these approaches we introduce a theoretical framework for analyzing interpretation,
which is applicable to interpretation of both human beings and computer models.
We give an overview of relevant computational approaches from various fields,
and discuss the most common and promising application areas.
The focus of this paper lies on interpretation of text and image data,
while many of the presented approaches are applicable to other types of data as well.

\end{abstract}


\end{frontmatter}



\section{Introduction}
Individual ways of interpretation play a major role in a variety of fields.
The philosophical positions scepticism, relativism and perspectivism all crucially involve the notion of points of view \cite{campos2015notion}, \ie different ways of interpretation.
Hermeneutics refers to a whole field that is concerned with how we interpret information
and commonly assumes that in order to make sense of things we need to relate them to our own life situation, which makes all interpretation something inherently personal (\eg see  \cite{zimmermann2015hermeneutics}).
Analyzing how we make sense of the world is pertinent to cognitive science, the research field concerned with studying the human mind.
Similarly, in psychology it has been argued that understanding each others' motivations is a key aspect of human social life \cite{forgas2005social}.
Even in a non-scientific context, everyday misunderstandings in communication offer a clear demonstration of both challenge and importance of correctly estimating what other people mean and anticipating how they would interpret our own behavior.

Nowadays, there are two developments that drastically impact our social life and motivate the need for computational methods with similar social abilities:
First, more and more communication is happening online \cite{poushter2016smartphone}.
Second, AI approaches have become much more ubiquitous.
This is especially prevalent online, where chatbots take part in discussions, recommendation algorithms suggest things we are likely to favor,
and search results are nicely ranked by yet another computer model.
In a broad sense, humans and computer models are all actors in a large communication network.
In many cases the goal of an AI approach is to learn about a certain way of interpretation.
This is most clearly visible in supervised approaches where the ground truth data serves as a proxy to the human perspective that is to be learned, which often involves estimating subjective qualities (\eg what a user will like, or even automatically mining opinions).
At the same time, as AI approaches become actors in communication and their automatic decisions become more and more influential in our everyday life,
we also have a motivation to understand them.
As approaches have grown considerably more complex over the years, this is not at all trivial.
However, since early 2018, with changes in European legislation (GDPR~\cite{EU-2016-679-GDPR}) there is now even a legal reason why many companies (and probably also researchers) should analyze how the developed models draw their conclusions: Whenever users are affected by automatic decisions, the users now have the legal right for an explanation of the decision in simple terms \cite{Goodman2016-EU-GDPR-Right-to-an-explanation}.
%
Yet another pragmatic motivation for understanding AI approaches stems from ever-growing amounts of data (``big data'') involved in digital activities such as posting comments, liking contents or browsing websites:
Due to the scale of user data, it has become extremely challenging to manually inspect even a fraction of the data.
Here, computers have a clear edge in terms of scalability,
and are valuable for processing all this information and thus making it more accessible to us,
potentially even by explaining its characteristics.

So we see that there are three important tasks, namely
enabling AI approaches to ``understand'' our view, understanding how AI agents see the world, and having computer models explain complex data to us.
It is clear that neither of these tasks is simple,
still, good progress has been made on all of them.
To name a few recent advances:
A lot of work was done on explaining how deep learning models work \cite{guidotti2018survey, SamITU18, MONTAVON20181, mythos, Palacio_2018_CVPR, mahendran2015understanding, mahendran2016visualizing},
which was even useful for helping us understand complex scientific data \cite{sturm2016interpretable, schutt2017quantum, alipanahi2015predicting}.
In case of data annotation, probabilistic methods have been proposed to merge annotator votes efficiently and simultaneously estimating annotator reliabilities \cite{branson2017lean, Horn_2018_CVPR}. 
However, despite related goals, approaches for interpretation analysis seem quite separated and we find an apparent lack of high-level bridges to connect them. In particular, recent surveys on explainability methods for machine learning  \cite{guidotti2018survey, SamITU18, MONTAVON20181, mythos} do not consider methods for comparing multiple ways of interpretation. 
Moreover, underlying concepts such as interpretation or understanding are often not defined properly (as \cite{mythos} explains for the concept interpretability),
which suggests the need for more rigorous formalism.

The main purpose of this paper is to connect various ideas and approaches, and put them into a coherent view. 
To this end, we introduce a theoretical framework, in which a perspective is 
represented by a function from input to meaning, called the interpretation function.
Interpretation analysis can then be understood mathematically as characterization of such an interpretation function.
We do a survey on approaches for this task with a focus on text and image inputs,
where we in particular find statistical methods, pattern mining, model-based approaches and visualization techniques to be of central relevance.
In addition to outlining methods for analyzing interpretations of a single model, this paper describes methods for comparing multiple perspectives.
We also unveil 
relations to the humanities, where it has a much longer tradition to look into characteristics of interpretation,
in the hope that this will contribute to more discussion between the disciplines.

We structure the paper as follows:
First, in Section~\ref{sec:framework} we will describe our theoretical framework and formally define interpretation, perspective and the task of interpretation analysis.
This is followed by general remarks about the task in Section~\ref{sec:methods}, where we comment on evaluation, ethics and input representation.
We will then look into approaches for the case of analyzing one individual perspective (Section~\ref{sec:single}).
To this end, we can make use of statistical methods, pattern mining, model-based approaches or visualization techniques (see overview in Table~\ref{tab:single_perspective}).
Comparisons between multiple perspectives will be handled in Section~\ref{sec:compare} and can be done under the use of three kinds of approaches (see also Figure~\ref{fig:comparing}).
We will see that two of these cases can mostly be reduced to single perspective analysis,
which makes the methods for analyzing relations between input and output of a single interpretation function the core of this paper.
In Section~\ref{sec:applications}, we outline five application fields, where ways of interpretation are analyzed by means of computational methods.
Finally, we close the paper with a few remarks on future work and ethical aspects (Section~\ref{sec:conclusion}).

\section{Theoretical Framework} \label{sec:framework}

Montavon et al. \cite{MONTAVON20181} define interpretation as a ``mapping of an abstract concept (\eg a predicted class) into a domain that the human can make sense of''.
We agree that this might work for the specific purpose of their analysis, but find this definition to be in conflict with intuition.
Most importantly, the definition does not include a large part of human interpretation,
which in general starts from something concrete (like an image or text) and ends up in something more abstract that we can broadly call meaning.
Hence, we keep the mapping part but remove the restrictions of the input and output domain while we introduce the notion of a \textit{bearer},
inspired by recent works in philosophy on defining perspectives \cite{campos2015notion, gutierrez2015subjective}:

\subsubsection*{Definition (Perspective, bearer, interpretation, interpretable)} 
We define a \textit{perspective} as a way of interpretation of some actor or group of actors $b$, which we call the \textit{bearer(s)} of the perspective.
Formally, we can represent a way of interpretation by a mapping from input to meaning, and call this mapping the \textit{interpretation function}~$f_b$ of $b$:
\begin{equation} \label{eq:perspective}
    f_b: I_b \rightarrow M_b \, ,
\end{equation}
where $I_b$ is the input domain 
and $M_b$ the output domain (set of potential meanings).
Any information $i$ is then called \textit{interpretable} by $b$
if and only if it is contained in the input domain of $b$'s interpretation function, \ie $i \in I_b$.

\subsubsection*{Examples}
1) Image classification of a machine learning model $m$ can be seen as interpretation process, where the interpretation function $f_m$ of the model maps from a set of images $I_m$ into a set of classes $M_m$.
2) An example for a human perspective would be the interpretation process of annotator $a$ from a set of tweets $I_a$ into \{sarcastic, not sarcastic\} when being asked to label tweets accordingly.
3) More complicated output domains are possible. For example, in case of an image autoencoder $e$ the latent representation can be modelled as interpretation of $e$.

\subsection{Role of the bearer}
We do not impose any particular requirements on the input or output domain, 
but we require that a perspective is adopted by some actor $b$ (\eg human being or computer model, existing or hypothetical), or group of actors.
In case a restriction is necessary, one can achieve this by limiting the set of possible bearers,
which naturally leads to restrictions on the input and output domains, as well as the form of possible interpretation functions.
For example, if $b$ is limited to be certain neural networks, both inputs and outputs are typically in tensor format.

Introducing a bearer in our definition of interpretation also paves the way to comparing ways of interpretation adopted by different bearers.
This can mean comparing perspectives of different people or perspectives of a single person under different circumstances (\eg happy vs sad).
In this way, our theoretic framework can be used to analyze the effects of contextual factors such as mood, geolocation or preceding events on interpretation.

Note that in the following, if only a single perspective is involved, we will usually not explicitly mention the bearer of the perspective and just use the symbol $f$ to refer to the interpretation function.

\subsection{Assumptions} \label{sec:assumptions}
For this paper, we assume that we do not have direct access to any interpretation function $f$,
but only have a list of inputs and their corresponding outputs.
In other words, we treat interpretation as a black-box, 
that is accessible only through a list of input-output pairs.
More precisely, if a single perspective is analyzed, the data is of the form $(d_0, f(d_0)), \ldots, (d_n,f(d_n))$, $n \in \mathbb{N}$.
Analogously, if multiple perspectives are involved, we assume the data to be of the form $(d_0, b_0, f_{b_0}(d_0)), \ldots, (d_n, b_n, f_{b_n}(d_n))$,
where $b_i$ describe the bearers of the respective perspectives.

The assumption that interpretation functions are not directly observable and perspectives are given indirectly as input-output pairs
enables us to more easily model interpretation of humans and AI approaches within the same framework.
This is another point that clearly distinguishes this survey from other overview papers related to explainability such as \cite{mythos, SamITU18, MONTAVON20181},
which assume that $f$ stems from a known machine learning model.

\subsection{Goals of interpretation analysis} \label{sec:goals}
\begin{figure}[t]
\begin{center}
   \includegraphics[width=0.75\linewidth]{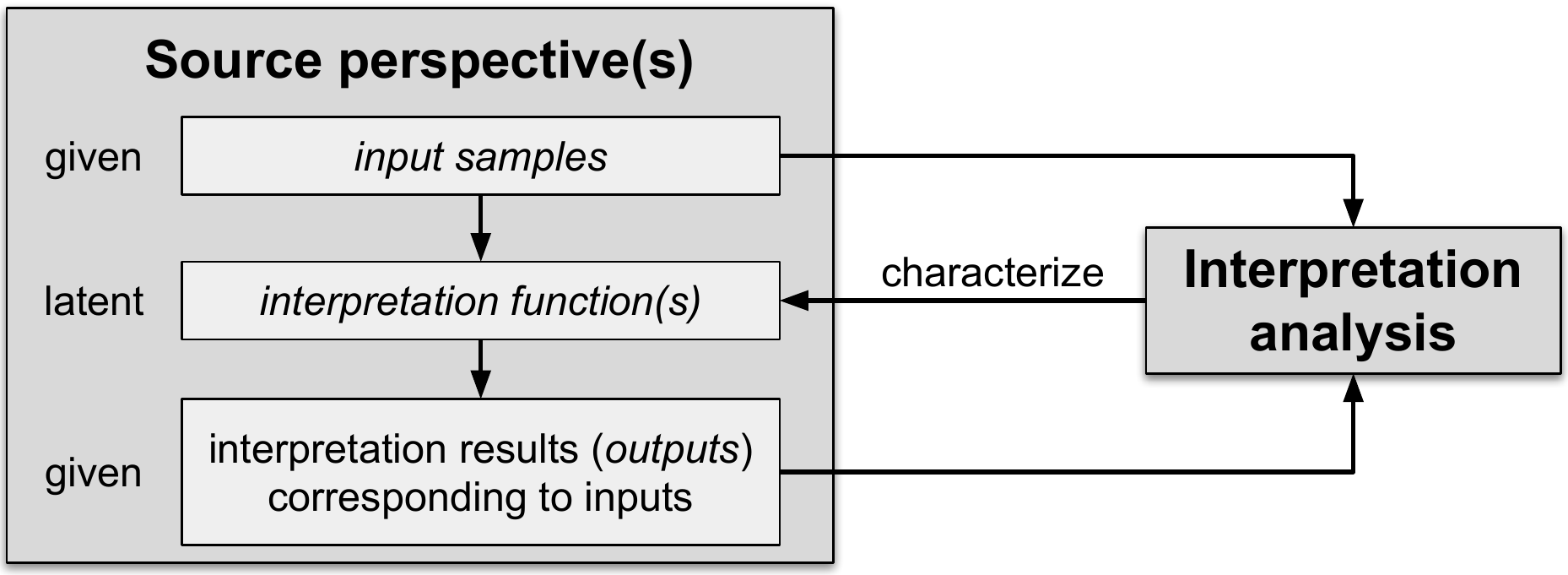}
\end{center}
   \caption{Interpretation analysis under the black-box assumption. The goal is to characterize interpretation from one or several perspectives,
   which can be human or artificial.
   Interpretation from each perspective is formally described as a mapping from information to meaning.
   For this paper, we assume that these functions are not directly accessible, but only indirectly via a list of inputs and associated outputs.}
\label{fig:interpretation_analysis}
\end{figure}

Overall, the main goal of interpretation analysis is to characterize interpretation functions.
(See Figure~\ref{fig:interpretation_analysis} for a schematic overview.)
Such a characterization can take different forms and be addressed in various ways,
depending in particular on whether the goal is to understand a single perspective (Section~\ref{sec:single}) or to compare several perspectives (Section~\ref{sec:compare}).

\begin{table*}[t]
  \begin{tabular}{|p{.20\textwidth}|p{.3\textwidth}|p{.4\textwidth}|}
    \hline
    {\bfseries Approach type} & {\bfseries Methods} & {\bfseries Outcomes}  \\
    \hline
    statistical methods 
    &
        correlation coefficients;
        hypothesis testing;
        CCA
    &
        measure of correlation, significance, canonical correlations
    \\
    \hline
    pattern mining 
    &
        association rule mining;
        emerging pattern mining;
        discriminative pattern mining 
    &
        association rules (implications), characteristic patterns
    \\
    \hline
    model-based approaches 
    &
        heatmapping;
        prototypes;
        globally understandable models;
        partially understandable models;
        ablation studies
    &
        model for approximating interpretation function, plus:
        explanations for individual decisions (heatmapping), characteristic inputs (prototypes),
        or approximate functional description of the function 
    \\
    \hline
    visualization techniques 
    &
        dimensionality reduction;
        example-based approaches;
        text summarization 
    &
        compression of the data, in form of plots, selected examples, or text summary
    \\
    \hline
  \end{tabular}
  \caption{Overview of approaches for single perspective analysis.}
  \label{tab:single_perspective}
\end{table*}

For analysis of a single perspective, we want to extract characteristic properties from a single function
in order to answer the question: ``What are the relations between features of the input and interpretation result?''
For example, which parts of the image make the classifier say that there is a dog in the image?

For comparing several perspectives, we are generally interested in discriminative characterization.
For example, we can ask ``For which kinds of inputs can we expect any difference between machine learning models A and B?''
or ``Which features of tweets characterize the set of tweets which annotator C labels as \textit{aggressive} while annotator D labels them as \textit{non aggressive}?''

\section{Computational Approaches} \label{sec:methods}
As we just saw in Section~\ref{sec:goals},
interpretation analysis in the proposed framework amounts to characterizing functions, interpretation functions to be more precise.
The general purpose of functions is to formally describe how one quantity (the output) depends on another quantity (the input).
Hence, at the very core of interpretation analysis (or analyzing and understanding any function for that matter)
we find the task of figuring out how outputs \textit{depend on} inputs.
And this is to be done based on a list of inputs and their corresponding outputs.
So we have already converted the conceptually challenging problem of interpretation analysis into a more graspable mathematical formulation,
which can be tackled with a variety of existing computational methods.
We have also discussed that the task takes on a slightly different touch
depending on whether we are analyzing one individual perspective or aim at comparing between multiple ones.
Before we go into detail on these approaches in Sections~\ref{sec:single} and \ref{sec:compare},
we will first discuss three general points that are relevant in all these cases,
namely evaluation, ethics and feature extraction.

\subsection{Evaluation} \label{sec:evaluation}
Natural 
questions to ask when being confronted with any large set of tools for a single task are:
Which one to choose? And on which grounds should one make such a decision?
So, how can we evaluate which method for interpretation analysis does the best job?

First of all, despite following the common goal of characterizing a single function in terms of relations between input and output,
the relevant approaches vary in terms of result format, but also
with respect to other properties such as reliability and expected data (type and amount).
This makes it difficult to directly compare all the approaches,
and indeed, a general automatic evaluation measure for interpretation analysis does not exist.
For several individual categories evaluative measures have been proposed (\eg see \cite{samek-tnnls17} for heatmapping),
but
in practice, quantifying usefulness of explanations largely remains an open issue and qualitative evaluation often becomes necessary.
This can mean that researchers manually inspect results and view examples for judging which method does the better job, 
or task someone else (\eg crowdworkers) with evaluating which method generates better explanations (\eg as in \cite{escalera2017chalearn}).
Another interesting option is mentioned in \cite{MONTAVON20181}, namely to look at simpler versions of the tasks where an optimal explanation can be specified and then compare the results to this explanation.

In general, we regard the following three criteria as important:
1) The results should be \textit{reliable}, which includes statistical significance and robustness.
2) The characterization should be simple to \textit{understand}.
3) The findings should \textit{cover} as much as possible of the \textit{variation} in the data that one wants to understand. (For a single perspective, explain variations in output in terms of input; for several perspectives, explain their differences.)
Note that these points are treated quite differently in the relevant fields.
Reliability is absolutely fundamental in statistics and still important in pattern mining,
but mentioned more rarely in model-based approaches. 
Understandability is a factor across the fields, but interestingly, the necessary background knowledge for correctly interpreting given explanations varies significantly.
Coverage of variation is often checked in statistics (coefficient of determination, $R^2$), quite central in pattern mining,
but harder to address in some of the model-based approaches (\eg how to measure to which degree output variation can be explained in terms of heatmaps or prototypes).

\subsection{Ethics}
We have just discussed various general criteria for judging the quality of analysis methods given a specific task.
However, if we zoom out and look at the big picture of interpretation analysis, it becomes clear that such analyses often have substantial ethical implications.
Thus, we find ethical considerations to be a crucial part of analysis, especially when dealing with human interpretation.

\subsubsection*{Analyzing human interpretation}
Research on human interpretation can help to improve user experiences
but also pave the way to ethically doubtful applications.
For example, better understanding how we interpret information can be used for ``computational propaganda'' \cite{woolley2016automation} and microtargeting, where people's personality traits are used to predict what kind of message is most likely to persuade them \cite{zuiderveen2018online}.

A nice starting point for ethical considerations can be the paper by Zook et al. \cite{zook2017ten}, which introduces ``ten simple rules for responsible big data research'', including many examples and pointers to further details.
We cite their ten rules here to provide a general idea, while we refer to their paper for details:
\begin{enumerate}
    \item ``Acknowledge that data are people and can do harm''
    \item ``Recognize that privacy is more than a binary value''
    \item ``Guard against the reidentification of your data''
    \item ``Practice ethical data sharing''
    \item ``Consider the strengths and limitations of your data; big does not automatically mean better''
    \item ``Debate the tough, ethical choices''
    \item ``Develop a code of conduct for your organization, research community, or industry''
    \item ``Design your data and systems for auditability''
    \item ``Engage with the broader consequences of data and analysis practices''
    \item ``Know when to break these rules''
\end{enumerate}

Importantly, these points should encourage thinking and discussing about ethical implications in the first place, but also make clear that ethics is not a simple matter.
In this context, we would like to recommend to not only discuss with researchers from computer science but form interdisciplinary collaborations.
This certainly does not automatically eliminate all potential negative consequences, but we believe that it does reduce the risk by safeguarding against very narrow perspectives.
Overall, we advice to start by asking questions such as ``Do we really want to analyze this aspect of interpretation?'' and ``Could such an analysis potentially do more harm than good?'' before jumping into technical details.

\subsubsection*{Interpretation analysis in the broader context of AI}
Recent advances in AI suggest a great potential for solving pressing social problems with the help of computer systems,
while building ethical AI requires us to wrestle with tough questions like ``Is this moral?'', ``Is it racist?'', ``Is it safe for everyone?'', ``Should we build it?''

Take for example the issue of predictive policing. There is a growing trend among law enforcement units globally (cities like Chicago, London and New York City) where 
big data and machine learning are used to predict potential criminals and surveil communication on social media platforms \cite{predictive_policing}.
Early on, this form of digital policing was touted as an innovative strategy for catching crime and violence before it happens \cite{mcclendon2015using}.
However, researchers and journalists have identified clear challenges that include: unconscious and implicit bias in the interpretation of language and images on social media that are deemed threatening \cite{falsified_data}, increased and disproportional surveillance of black and brown communities \cite{freelon2016beyond}, increased arrest of individuals who pose little threat and missed predications of white perpetrators of crime and violence \cite{doi:10.1177/2056305117733344}. 

One practical response is the creation of critical and diverse partnerships between computer scientists, community members and law enforcement that
reviews interpretation of images and text across race, ethnicity and culture,
analyzes system outputs for racial and cultural sensitivity,
and considers the implications of AI tools for community well-being and safety.
Within such an environment, we see great potential for interpretation analysis techniques by using them for revealing problematic biases in training data or AI systems.

\subsection{Feature extraction}
Lastly, the third generally applicable point is feature extraction.
Here, and in most of machine learning, we face a situation similar to that in correlational studies in psychology \cite{levitin2002},
where the data is already there and we need to answer:
What is the kind of input ``parts'' we want to consider for checking dependencies with the output?

First of all, many of the approaches we will discuss cannot be expected to reveal interesting findings when applied to low-level input features such as individual pixels or sequences of characters.
For example, if the color of any individual pixel of an image correlates significantly with a classifier output for ``dog'', then this is hard to make sense of and has a high chance of being a statistical artifact or a flaw in the training data.
This is per se not specific to interpretation analysis and especially in applied machine learning feature engineering (\ie finding suitable features) remains a key part \cite{domingos2012few} despite the efforts of the deep learning community for end to end learning.
This process generally requires expertise, since the features need to be appropriate for the final method, the data at hand, and the overall purpose of analysis.
It is in the last of these parts, purpose of analysis, where we find a considerable difference between standard machine learning and interpretation analysis.
Most of the time, in machine learning the features are meant to serve the purpose of building a prediction model that is reliable (\ie does not overfit) and has good predictive power.
In case of interpretation analysis, we have seen both of these criteria in similar forms (predictive power corresponding to coverage of variance),
but in addition require that results should be understandable (see Section~\ref{sec:evaluation}).

This leads to some features such as intermediate activations of a Convolutional Neural Network (CNN) being less straightforward to use.
After all, if for instance the 10th neuron of the penultimate layer from a VGG network \cite{simonyan2014very} was found to correlate with another image classifier's positive decision for the dog class, wouldn't this tell us more about VGG-based embeddings than about how the classifier interprets images?

Still, when deciding on which features to use, one should definitely be inspired by existing approaches on feature extraction, 
and some of the simpler common features (\eg bag of words, occurrences of specific n-grams, color histograms, bag of visual words) can be useful for analyzing interpretation.
Finally, in interpretation analysis it happens at times that features are implied by the research goal.
For example, if one wants to analyze whether a visual sentiment classifier prefers cats over dogs, cat and dog presence are suitable features.
Overall, finding the right features is a complex topic, in part because the understandability criterion is hard to formalize and its implications depend on the type of approach that is used later.
Hence, we will mention approach-specific examples in some of the following sections (\ref{sec:pattern_mining} and \ref{sec:model_based}).

\section{Input-output dependencies} \label{sec:single}
We now discuss computational approaches for understanding a single perspective.
Typical examples would be to
analyze which words in a social media post correlate with large numbers of likes (as a form of positive interpretation from the group of viewers),
or to analyze an image classifier based on a list of image-classification results for identifying which patches of images are most relevant for a particular result.

The formal context can be summed up as follows (also see Section~\ref{sec:framework}):
The perspective is described by an interpretation function $f: I \rightarrow M$ of interest.
This function is not given directly, so the goal of analysis is to determine relations between the function's input and output
based on a list of input-output pairs $(d_0, f(d_0)), (d_1, f(d_1)), \ldots, (d_n, f(d_n))$,
where $n \in \mathbb{N}$ and $d_i \in I$ for all $i$.
We are primarily interested in cases where $I$ consists of language data, images, or feature vectors thereof.
The output domain $M$ is assumed to contain feature vectors of fixed dimension.

For such a task we have several types of approaches from various well-established fields at our disposal,
which we will now discuss.
We group these approaches together into sections which roughly correspond to research fields (statistical methods, pattern mining, model-based, visualization).
In each section, we then organize techniques by their outcome or the goals they are aiming at. Each section is concluded with remarks on the usage of the respective type of analysis approach. (See overview in Table~\ref{tab:single_perspective}.)
For several of these, we will use hypothetical user preference data for illustration.
This data can be found in Table~\ref{tab:pattern_mining} and corresponds to a simple interpretation from a 3-D feature space into the binary space of like/dislike.
\begin{table}[h]
  \begin{center}
  \begin{tabular}{|c||c|c|c||c|}
    \hline
    {\bfseries Image ID} & {\bfseries Nudity} & {\bfseries Humor} & {\bfseries Explosions} & {\bfseries Like}  \\
    \hline
    0 & 0 & 1 & 0 & 0 \\
    1 & 1 & 0 & 1 & 1 \\
    2 & 0 & 1 & 1 & 1 \\
    3 & 1 & 1 & 0 & 0 \\
    4 & 1 & 0 & 0 & 1 \\
    5 & 1 & 1 & 1 & 0 \\
    \hline
  \end{tabular}
  \caption{Hypothetical image preference data of a single user. The three columns in the middle describe features of the image, while the last column describes the type of user reaction which corresponds to an interpretation result (assuming the user has the option to \eg either vote up or not).}
  \label{tab:pattern_mining}
  \end{center}
\end{table}

\subsection{Statistical methods} \label{sec:statistics}

One way to analyze relations between two quantities is to test for statistical dependencies between them.
We can treat both input and output as values of (composed) random variables $X$ and $Y$ respectively,
and then test whether individual dimensions $X_i$ of $X$ and $Y_j$ of $Y$ are \textit{statistically dependent}.
Formally, such a dependency is given if for any sets of possible values $A$ and $B$,
the two events $X_i \in A$ and $Y_j \in B$ are not independent, i.e., 
$P(Y_j \in B \mid X_i \in A) \not= P(Y_j \in B)$.
In other words this means that information about the value of $X_i$ can give us any information about the value of $Y_j$.
In our toy example (Table~\ref{tab:pattern_mining}), we could check if the image preference of the user statistically depends on
whether the image contains nudity, humor or explosions.
This can be done either by quantifying \textit{correlation} of user preference to individual input features of interest (\eg user preference to presence of explosions in the image) or by \textit{testing hypotheses} (\eg ``Is the user more likely to like an image if there is nudity?'').

Instead of analyzing the relation between individual input features $X_i$ to output components $Y_j$, by applying \textit{Canonical Correlation Analysis} it is also possible to find out which combination of features correlate with which combinations of output components.

\subsubsection*{Correlation coefficients}
In its broadest sense, correlation refers to 
any statistical dependency between two random variables.
More specifically, there exist several ways of calculating \textit{correlation coefficients}, each one of them designed
to measure the strength of a particular kind of statistical dependency.
The most common candidates are Pearson's correlation coefficient \cite{pearson1895note}, which measures linear dependence between two continuous random variables, and Spearman's rank correlation coefficient \cite{well2003research}, which measures how well the relationship between the two variables can be described by a monotonic function.
Both of these coefficients are fairly simple to interpret, however, it shall be noted that a Pearson or Spearman coefficient of $0$
does \textit{not} imply the absence of any statistical dependency between the variables.
For example, for $X$ uniformly distributed on $[-3,3]$ the random variables $X$ and $X^2$ have Pearson and Spearman correlation $0$ but are far from independent.
There exist other correlation measures, which are able to capture more complex statistical dependencies but are typically harder to interpret.
These include distance correlation introduced in \cite{szekely2007measuring}, which is $0$ only if the tested variables are independent.
Specific choices should be made based on the properties of the tested variables (distributions they follow) and the questions one is trying to answer with the analysis.

\subsubsection*{Statistical significance}
Correlation coefficients mainly measure the degree of a certain statistical dependency, but one should also check reliability of the findings by testing
whether the dependency is \textit{statistically significant}.
This can be done based on \textit{hypothesis testing} for estimating how likely it is that the true correlation is $0$ (in a two-sided test, or $\le 0$ or $\ge 0$ in one-sided tests)
and the observed correlation value is due to noise. 
Another option is to calculate \textit{confidence intervals} for the coefficients, for which a variety of methods have been proposed (\eg see  \cite{ruscio2008constructing} for Spearman correlation).

Note that, coming directly from the definition of statistical dependency,
we can also estimate confidence intervals for both the expected value of $Y_j$ and the expected value of $Y_j$ given a particular value $x$ of $X_i$.
If these confidence intervals do not overlap, this means that there is a significant difference between $E(Y_j)$ and $E(Y_j\mid X_i=x)$,
\ie $X_i$ attaining value $x$ significantly affects the expected value of the output $Y_j$.
It shall be mentioned that overlapping confidence intervals do \textit{not} imply that there is no significant difference \cite{knezevic2008overlapping}.

\subsubsection*{Canonical Correlation Analysis (CCA)}
Another set of statistical methods for analyzing the relation between two sets of variables (such as input and output variables in our case) is consituted by \textit{Canonical Correlation Analysis}, or short CCA.
The original CCA approach \cite{hotelling1935most,hotelling1936relations} aims at finding linear relations between a matrix of input observations and a matrix of output observations.
That is, if we are given a matrix $M_X$ with $m$ input features of $n$ items and a matrix $M_Y$ with $o$ output values for the same $n$ items as columns, the first objective is to find two vectors $z^1_X$ and $z^1_Y$ that map $M_X$ and $M_Y$ on the $n$-dimensional unit ball such that the cosine similarity between $M_X z^1_X$ and $M_Y z^1_Y$ is maximized (\ie the transformations of $M_X$ and $M_Y$ point in a similar direction). Iteratively, further vectors $z^i_X$ and $z^i_Y$ are calculated under the additional constraint that each $z^i_X$ (and $z^i_Y$ resp.) must be orthogonal to all previous vectors $z^j_X$ ($z^j_Y$ resp.), for all $j=1,\ldots,i-1$ (see \cite{uurtio2018tutorial}).

Different from computing correlation coefficients between individual features, CCA belongs to multivariate statistics, and returns correlations between combinations of features.
For example, in case of our toy example from Table~\ref{tab:pattern_mining}, CCA gives us a result of the form
$$ z^1_X = (0.43, 0.85, -0.30)^T, \quad z^1_Y = (-1) \,,$$
indicating that a linear combination of $0.43$ nudity, $0.85$ humor and $-0.3$ explosions correlate maximally \textit{negatively} with the user's preference.
Statistical significance of CCA results can be checked by appyling Barlett's sequential test procedure \cite{bartlett1941statistical}.

It is worth mentioning that CCA falls under dimensionality reduction techniques as well \cite{cunningham2015linear}, a set of techniques which we will discuss below in Section~\ref{sec:visualization}.
Furthermore, various modifications of CCA have been suggested.
These include kernel-based \cite{lai2000kernel} and neural network based methods \cite{lai1999neural, andrew2013deep} for finding non-linear relations, as well as techniques that aim at improving interpretability of the discovered relations by enforcing sparsity on non-zero coefficients \cite{parkhomenko2007genome,waaijenborg2008quantifying,witten2009penalized,hardoon2011sparse}. 
For further details we refer to the comprehensive and recent tutorial on CCA by Uurtio et al. \cite{uurtio2018tutorial}.

\subsubsection*{Remark on causality}
Intuitively, we might wish to understand which features of the input \textit{cause} a certain response.
For example, an analysis of user preferences might ultimately aim at helping to design new contents by pointing at specific features that are linked to positive user reactions and thus are suggested to be incorporated.
However, all methods we discussed try to figure out statistical dependence (correlation), which does not imply causation.
In fact, causal assumptions can generally only be verified if experimental control is exerted \cite{pearl2009}.
In the general case described in this paper, the possibility for collecting additional data while manipulating parts of the input cannot be guaranteed.
It shall be mentioned that for the case of analyzing given AI approaches,
this possibility is likely to be given and there are some recent attempts in computer science to address causality (\eg \cite{NIPS2008_3548, lopez2015towards, zhang2009identifiability, shimizu2011directlingam, stegle2010probabilistic}).
We believe that this direction should be further explored for interpretation analysis in future work,
and refer the interested reader also to the paper of Pearl \cite{pearl2009} for a solid overview on causal inference in statistics.

\subsubsection*{Usage}
Individual correlation coefficients are simple to understand,
the methods for computing them are transparent and concrete statements about reliability can be made.
Overall, correlation coefficients provide a robust way of quantifying the role of individual features as long as the feature space is not too high dimensional.
On the downside, results crucially depend on selecting the right input and output features for analysis, which can be very challenging to do.
This problem is less severe in CCA, which can pick up more complex dependencies.
However, in comparison to correlation coefficients, canonical correlations tend to be harder to make sense of.
An important advantage of statistical methods is that they allow for significance testing, which is necessary if specific claims in the form of hypotheses are to be tested rigorously.

\subsection{Pattern mining} \label{sec:pattern_mining}
The general goal of pattern mining is to find characteristic patterns in the data. 
What exactly constitutes a pattern varies, but they often take on the the forms of association rules, emerging patterns or visual patches, as will be described in the following.

\subsubsection*{Association rule mining}

Association rule mining has a long tradition in pattern mining \cite{piateski1991knowledge, agrawal1993mining}.
In particular, it is often used for web personalization where it is applied to usage data \cite{eirinaki2003web, mobasher2000automatic, mobasher2001effective}.
In its original form \cite{agrawal1993mining} it can be used to process a list of binary vectors and find implications of the form
``if an image contains nudity and humor, then in 50\% of cases the image also contains explosions'' (using hypothetical data from Table~\ref{tab:pattern_mining}).

Let $T = \{b_1, \ldots, b_n\}$ be a multi-set of $n$ transactions over $k$ items represented as binary vectors with $b_i \in \mathbb{B}^k$, $n,k \in \mathbb{N}$.
An association rule can formally be defined as implication of the form $X \Rightarrow j$,
where $X \subseteq \{0,\ldots,k\}$ is a set of indices called the antecedent of the rule,
and $j \in \{0,\ldots,k\} \setminus X$ is a single index (not included in $X$) called the consequent of the rule.
The \textit{support} of a set of indices $X$ can then be defined as the relative amount of transactions containing all items in $X$, and the \textit{confidence} of a rule $X \Rightarrow j$ as the relative support of the rule's antecedent and consequent over the support of its antecedent (see \cite{agrawal1993mining}): 
\begin{equation}
    supp(X) := \frac{| \{ b_i \in T \mid b_{i,j} = 1, \forall j \in X \} |}{|T|}
\end{equation}
\begin{equation}
    conf(X \Rightarrow j) := \frac{supp(X \cup \{j\})}{supp(X)}
\end{equation}

Another important measure \textit{lift} \cite{brin1997dynamic},
describes the ratio of the observed support for a rule to the support that would be expected if antecedent and consequent were independent.
Confidence, support, other measures such as lift, and given potential constraints (\eg only considering rules with specific $j$),
can all serve as criteria for filtering possible rules. 
Association rules are often computed based on the apriori \cite{agrawal1994fast} or frequent pattern tree \cite{han2000mining} algorithms (see \eg the survey \cite{zhao2003association}).

For interpretation analysis, we are interested in so-called classification rules, \ie rules that have a subset of the input as antecedent and a subset of the output as consequent \cite{agrawal1994fast}.
So in our hypothetical example (Table~\ref{tab:pattern_mining}), we would try to find rules of the form
``if an image contains explosions, then the user likes it in 2/3 of cases.''
Such a way of modeling is for instance adopted in \cite{quack2007efficient}, where association rule mining is used for finding class-discriminative features in images.
In their approach, a binary class membership entry is appended to all vectors and only rules with this particular index as consequent are considered.

\subsubsection*{Emerging pattern mining}
The problem of emerging pattern mining was introduced in \cite{dong1999efficient}, originally for capturing trends in time-stamped databases.
It is similar to association rule mining, but uses the notion \textit{growth rate} to measure how support for a pattern (set of indices) differs between sets.
So broadly speaking, the goal of emerging pattern mining is to find differences in patterns across multiple sets.
Soon after the task was introduced, it has been used for classification purposes \cite{dong1999caep, li2000instance}, where emerging patterns are meant to capture characteristic differences between classes.
To this end, input samples are partitioned based on the associated output values and found patterns used to discriminate between the resulting partitions.
It is in this sense that this approach can directly be used for interpretation analysis.
Coming back to our toy example of Table~\ref{tab:pattern_mining}, following an emerging pattern mining approach we would ask,
which are the combinations of nudity, humor and explosions that are comparatively more frequent in images the user likes/dislikes.
Note that the survey of Novak et al. \cite{novak2009supervised} puts emerging pattern mining under the umbrella term supervised descriptive rule discovery,
together with contrast set mining and subgroup mining.
Another useful resource is the recent survey of \cite{garciavico}.

\subsubsection*{Visual pattern mining}
There are several image-specific approaches worth mentioning.
In \cite{rematas2015dataset}, Rematas et al. use standard data mining terminology to formulate the problem of finding characteristic visual patches from a given image collection,
which they also put into a graph for navigation through the image collection.
The publications \cite{li2015mid, Li2017} use association rule mining on mid-level CNN features, and call this combination mid-level deep pattern mining. 

Note that sometimes the notion ``parts'' is used for referring to something comparable to visual patterns.
For example, \cite{parizi2014automatic} describes how to automatically discover discriminative parts for the purpose of image classification.
Visual pattern mining was also applied in \cite{cruz2011visual},
by using a bag-of-features representation (also known as bag-of-visual-words) \cite{csurka2004visual}
and selecting representative and discriminative local features based on Peng's method for feature selection \cite{peng2005feature}.
The recently proposed PatternNet \cite{Li:2018:PVP:3206025.3206039} introduces a CNN that directly learns discriminative visual patterns.
(As such, some of these approaches could as well be put into the model-based category described in the next section.)

\subsubsection*{Usage}
Pattern mining approaches are conceptually similar to the statistical methods discussed above, as they discover relations between input and output features.
The crucial difference is that in pattern mining approaches, these relations are described in different formats, which are designed to be intuitively understandable and can take the form of rules, discriminative patterns or characteristic visual patches.
However, understanding can be hard for more complex patterns (\eg very long rules) and while pattern mining techniques still include measures for reliability of the findings, there might be a high risk of ending up with many false alarms, since the space of possible patterns can be huge \cite{webb2007discovering}.
Also note that many pattern mining techniques operate on binary data, so it might become necessary to first convert the data.
In the above-mentioned paper \cite{quack2007efficient}, this is done for example by choosing a bag-of-features image representation.
An example of an adaptation of pattern mining to textual data can be found in \cite{zhong2012effective}.

\subsection{Model-based approaches} \label{sec:model_based}
Even though the perspective of interest is considered to be a black-box in this paper,
it is still possible to build another model to approximate the interpretation function based on the given input-output pairs.
Successively, this trained model can be analyzed in the hope to reveal information about data dependencies that the original black-box might also rely on.
In the example of our toy data (Table~\ref{tab:pattern_mining}), we would first train a computational model to predict like/dislike from the input features nudity, humor and explosions,
and successively analyze the trained model for dependencies between both parts.

We will discuss four kinds of model-based approaches which each focus on different aspects of analysis:
First, \textit{heatmapping} techniques aim at visually explaining decisions for individual items (\eg ``Which image features are likely to make the user like an individual image?'').
Second, \textit{prototype} approaches compute characteristic inputs for the different output classes (\eg ``How does a typical image look like which the user favors?'').
Third, globally or partially \textit{understandable models} can be used to approximate the perspective in order to obtain a more holistic understanding of how interpretation works.
Finally, in \textit{ablation studies}, the role of individual input features or model components is analyzed by removing them.

We do not go into too much detail for heatmapping and prototype methods because there are other survey papers such as \cite{samek-tnnls17, MONTAVON20181}
which give an excellent overview for most of these approaches (in a non-black-box set-up).
Similarly, \cite{guidotti2018survey} contains a comprehensive treatment of globally understandable models.
Partially understandable models and ablation studies are less frequently mentioned in the context of explainability methods in machine learning research.

\subsubsection*{Heatmapping} 
\begin{figure}[ht]
\begin{center}
\subfloat[Image]{\includegraphics[width=0.48\columnwidth]{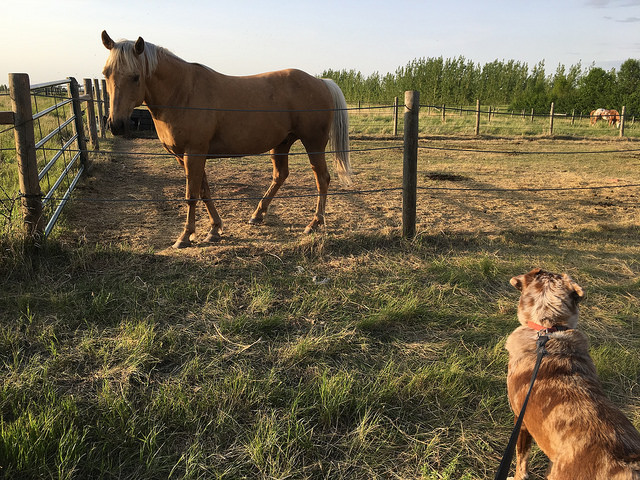}}
\hfill
\subfloat[Heatmap]{\includegraphics[width=0.48\columnwidth]{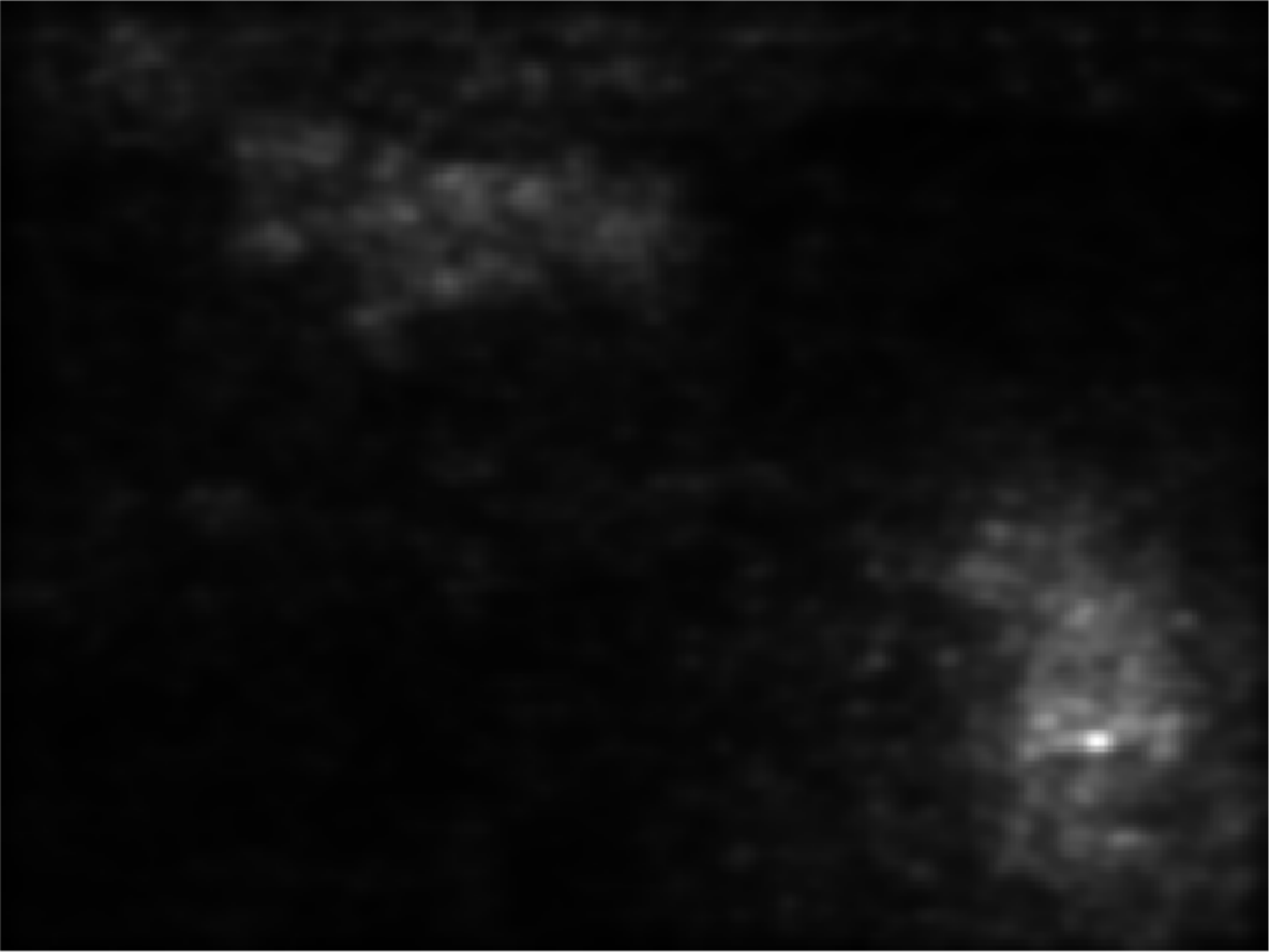}}
\end{center}
   \caption{Example of a heatmap computed from an inception-v3 \cite{inceptionv3} network which was trained for image classification on the imagenet dataset \cite{imagenet}. The top predicted class for this image was ``malinois'' (particular dog breed). The heatmap shows absolute values of input gradients, which serve as visual explanation of the classification result for this particular image. Smoothing and logscale have been applied to the gradients for illustration purposes. Image ``Hey Big Dog!'' by Alan Levine (\url{https://www.flickr.com/photos/cogdog/41916073004}, public domain).}
\label{fig:heatmap_examples}
\end{figure}

In the context of analyzing machine learning models, a \textit{heatmap} refers to an explanation of the model's decision for a particular sample in terms of the input,
indicating visually which parts of the input are relevant (positively or negatively) for the decision.
For an example of a heatmap, see Figure~\ref{fig:heatmap_examples}.
Heatmapping techniques can broadly be classified as methods for computing \textit{saliency maps} and \textit{relevance methods}.

A common method is to calculate \textit{saliency maps} \cite{baehrens2010explain, rasmussen2012visualization, simonyan2013deep} based on sensitivity analysis \cite{zurada1994sensitivity, sung1998ranking, saltelli2008global},
\ie the gradients on the model's input are used for estimating how sensitive the model is to changes in the individual input components.
A related approach is prediction difference analysis \cite{zintgraf2017},
which is still based on sensitivity analysis 
but uses local regularization in order to obtain visualizations that are easier to interpret.
Saliency maps are simple to calculate for neural networks by means of backpropagation \cite{rumelhart1986learning},
but on the downside, resulting heatmaps have been shown to be unreliable in certain cases \cite{kindermans2017reliability}.
Also, sensitivities to input components is typically not exactly what we want to find out,
because they only tell us how the input could be changed to make it belong more or less to a certain class
instead of explaining which parts of the input actually make it belong to a class.

The latter can be achieved with \textit{relevance methods} within the theoretical framework of Taylor decomposition \cite{bazen2013taylor}.
In \cite{bach2015pixel}, Bach et al. adapt Taylor decomposition to neural networks and introduce layer-wise relevance propagation (LRP),
which makes use of the network's architecture to propagate relevance backwards through the network for obtaining a heatmap.
The backward propagation rule they derive takes two hyperparameters and for one particular combination,
simplifies into a rule that is interpretable as deep Taylor decomposition \cite{montavon2017explaining}.
Other backprop techniques have been proposed for computing heatmaps for neural networks,
including Deconvolution \cite{zeiler2014visualizing}, Guided Backprop \cite{springenberg2014striving}, 
Class Activation Mapping \cite{zhou2016learning},
PatternAttribution \cite{kindermans2017learning} and PatternLRP \cite{kindermans2017patternnet}. 

\subsubsection*{Prototypes}
Another way of visualizing what the model has learned is to calculate inputs that serve as prototypes for the individual classes.
For example, a neural network trained on the MNIST dataset to recognize the digits 0-9 can be used to obtain a ``typical'' image for the digit 5.
Prototypes can be calculated within the analysis framework of \textit{activation maximization} \cite{berkes2006analysis, erhan2009visualizing}.
Essentially, finding prototypes amounts to solving the optimization problem of finding an input that maximizes a certain component of the output
(\eg an image that is interpreted by the model as being maximally dog-like).
Without any additional restrictions, the resulting prototypes tend to be unnatural \cite{simonyan2013deep},
which is why various regularization methods have been proposed \cite{mahendran2015understanding, nguyen2017plug, nguyen2016synthesizing}.
%
For neural networks in particular, there are numerous efforts on visualizing what particular neurons or neuron layers have learned (\eg \cite{krizhevsky2012imagenet, zeiler2014visualizing, yosinski2015understanding})
which can also be seen as prototype approaches.
An interesting non-prototype (but still related) approach is the one of \cite{bau2017network, zhou2018interpreting},
where hidden unit activations are related to a binary segmentation task of the input for a given list of semantic concepts,
in order to analyze semantics of individual hidden units.

\subsubsection*{Globally understandable models}
Depending on the complexity of the data, it is possible to train a model that approximates the whole interpretation function in an understandable way.
The most common candidates for such globally understandable models are linear models, decision trees and rules \cite{guidotti2018survey}.

Linear models assign a weight to each feature, which provides a direct measure of the feature's importance in terms of sign and magnitude. 
Especially in the social sciences it is common practice to use analysis of variance (ANOVA) \cite{fisher1921probable} for analyzing experimental data.
ANOVA is considered to be a special case of linear regression \cite{montgomery2017design}.

Decision trees are tree-like graphs, where internal nodes represent tests on input features and the leaf nodes represent a certain output.
Decision trees have been used extensively since the early days of machine learning (see \eg \cite{safavian1991survey} and \cite{murthy1998automatic}).
Similar to decision trees, decision sets \cite{Lakkaraju:2016:IDS:2939672.2939874}
or decision lists \cite{Rivest1987} can be compiled from data.
Note that conceptually many of these approaches are very closely related to association rule mining (which we discussed in Section~\ref{sec:pattern_mining}).
In fact, decision trees can be converted to sets of decision rules \cite{quinlan1987generating}.

An interesting option that does not fall into any of these standard categories for understandable models are hypothesis-based models, which are common in the field of computational psychiatry \cite{adams2016computational}.
There, conflicting hypotheses are implemented as computational models and fit on the given data to find which of the models (and therefore hypothesis) is better suited for explaining human information processing.

\subsubsection*{Partially understandable models}
If the data is too complex for globally understandable models to fit properly,
partially understandable models can be an appropriate compromise between understandability and prediction power.
We discuss two ways of achieving this compromise: One is by breaking the problem down into more accessible steps in \textit{pipeline approaches},
the other is to incorporate specific \textit{structural components} into architectures that can be understood intuitively (\eg explicit attention mechanism).

In pipeline approaches, specific mid-level features can be used to simplify understanding of the model's output.
For example, \cite{VSO, chen2014deepsentibank} take the detour of recognizing adjective-noun combinations in images for the task of visual sentiment detection,
and \cite{blandfort2018multimodal} propose a list of visual concepts to be used as intermediate features for classifying multimodal tweets of presumably gang-associated youth.
Explicit attention mechanisms were mentioned above as one way to include understandable components into architectures.
Such attention mechanisms are frequently used in machine translation \cite{luong2015effective},
are a key component of memory networks \cite{sukhbaatar2015end, weston2014memorynet},
and have been used for tasks such as image captioning as well \cite{xu2015show}.
A related approach is that of \cite{zhang2018interpretable}, which explains how to modify CNN architectures such that learned filters are more semantically meaningful and understandable.

\subsubsection*{Ablation studies}
The principle of ablation studies is to gain understanding of the role of a system's components by analyzing how the overall system changes if the component is removed.
Historically, in neuroscience many early insights about functionality of individual brain regions were obtained by examining changes resulting from brain damage in particular areas \cite{kandel2000principles}.
In computer science, ablation studies have been adopted for quantifying the importance of model components \cite{matuszek2012joint},
which for example can be used for model verification or reduction. 

For interpretation analysis, ablation can be a useful tool when applied at the input level to address two points:
First, which parts of the input are necessary for approximating the perspective of interest?
If prediction performance drops drastically after removing a certain feature from the input, the feature was important for learning.
This principle is frequently made use of in NLP for analyzing the role of features for prediction (\eg for identifying hate speech \cite{schmidt2017survey}).
Second, when having trained a model for perspective approximation, one might want to verify that the model does not use any parts of the input which it should not use
(\eg because they might be known not to be used by the original interpretation function).
For example, \cite{ablation2017fish} uses an ablation study where they mask the foreground to confirm that the classifier does not cheat by predicting from background properties.

\subsubsection*{Usage}
In principle, model-based approaches can be used to learn complex dependencies, and heatmapping can explain decisions in individual cases,
even when training models directly on pixel data \cite{montavon2017explaining} or word sequences \cite{arras2017}.
Heatmapping has been applied together with several other features too, such as bag of visual words \cite{bach2015pixel} and fisher vectors \cite{lapuschkin-cvpr16}.
Such model-based explanations were found to be useful
in many publications (\eg \cite{schutt2017quantum, sturm2016interpretable, zintgraf2017}).
Zhou et al. \cite{zhou2016learning} also show how a network can learn to localize objects with decent performance without any bounding box labels.

Still, in general it is not clear which properties of the original perspective carry over to the trained model when fitting it on a given list of inputs and outputs,
and to the best of our knowledge, there is no extensive study analyzing the transfer of various functional properties. 
Indeed, publications dealing with adversarial noise (\eg \cite{papernot2016limitations, papernot2017practical, Folz2018AdversarialNoiseDefenseS2SNet}) show how convolutional neural networks are typically sensitive to things which humans are not \cite{szegedy2013intriguing, nguyen2015deep,Palacio_2018_CVPR},
despite being trained on large amounts of humanly annotated data and convolutional neural networks originally being inspired by human vision \cite{lecun1995convolutional}.
This gives reason for caution when making claims about the original function based on analyzing its approximation,
especially also for complex approximation methods such as deep neural networks.
If the models are simpler and do not have the capacity for picking up on any complex noise, some of these issues can be ruled out and the approach becomes closer to statistical testing.
Other partial remedies are to rely on pipeline approaches, where individual steps can be verified separately,
or make use of ablation studies to rule out certain unwanted properties.
Still, one should not confuse the trained model with the original perspective of interest,
and be aware that there often is a remaining risk that findings are unreliable or misleading.

\subsection{Visualization techniques} \label{sec:visualization}
In the following, we describe visualization techniques in a very broad sense as methods
to obtain a condensed representation of some given data.
This representation can take various forms:
In \textit{dimensionality reduction}, the data is transformed into a lower-dimensional space such that it can be plotted.
Other methods stay closer to the original type of data and rather reduce the amount of information in different ways.
These include extraction of \textit{examples}, reducing the amount of information by topic modeling, or automatic \textit{summarization}.

\subsubsection*{Dimensionality reduction}
Dimensionality reduction can be useful for visualizing almost any kind of data by reducing the data dimension,
such that it can then be plotted and manually inspected.
There are many different kinds of dimensionality reduction and several surveys have been made on the topic \cite{sorzano2014survey, gisbrecht2015data, cunningham2015linear}.
Here, we outline a few popular cases that are especially relevant for interpretation analysis.
Linear dimensionality reduction refers to methods that linearly transform the original input space, \ie they describe how to find a matrix that is multiplied to all inputs for projecting them into a smaller space (see \cite{cunningham2015linear}).
Popular methods that fall into this category are Principal Component Analysis (PCA) \cite{pearson1901liii}, Linear Discriminant Analysis (LDA) (\eg \cite{mclachlan2004discriminant}), and Canonical Correlation Analysis (CCA) \cite{hotelling1936relations},
which all compute orthogonal matrices for the transformation.
LDA uses associated class labels and transforms the input space such that after transformation the separation between the classes is maximized.
This is closely related to linear regression, which can be seen as another linear dimensionality reduction technique that does not use an orthogonality constraint.
An interesting property of PCA is that after transformation the components are linearly uncorrelated or, in other words, the data is factorized into independent components.
Other popular factorization methods include Factor Analysis (FA) \cite{spearman1904general}, which is widely used in psychology \cite{fabrigar1999evaluating}, 
for example to become aware of patterns in questionnaire items \cite{briggs1986role}.

Linear dimensionality reduction with orthogonal matrices can be especially helpful for getting a rough idea of the data's structure,
since they do not exaggerate relations between data points (see \cite{cunningham2015linear}).
Projections of non-linear transformation techniques can be harder to interpret since geometric properties like distances in the original space are generally not preserved.
Still, such techniques can be useful for looking at specific properties of the data, and there are a few non-linear transformation techniques that deserve mentioning:
t-SNE \cite{maaten2008visualizing} is a probabilistic method that embeds samples into a low-dimensional space such that similar samples are likely to be embedded to nearby points and dissimilar object to distant points.
Another non-linear reduction technique is to train an autoencoder \cite{bourlard1988auto} to compress the original data into a smaller latent encoding.
The benefit of autoencoders is that they can be combined with additional loss functions for enforcing other properties on these encodings,
such as following a certain distribution \cite{makhzani2015adversarial} or using specific positions to encode certain semantic properties \cite{hu2017toward}.

\subsubsection*{Example-based approaches}
The idea behind example-based approaches is that even for large collections, looking at characteristic examples
can be useful to automatically form a holistic understanding of the collection.
The crux herein is to select the right examples (and know how many are necessary),
for which various approaches exist.

A simple and yet useful method is to randomly select a few samples for manual inspection.
This cannot be expected to lead to a full understanding of the sample collection but helps to form an initial feeling for the data.
One issue is the possibility that by chance odd samples are drawn,
which are included in the data, but exhibit certain unexpected properties.
Obtaining such abnormal examples can also be done on purpose, which relates to a common task called anomaly detection (see \eg \cite{chandola2009anomaly}). 
Anomalies can for example help to become aware of problems with the data (\eg broken entries),
but can also be of particular relevance when working with methods that are sensitive to statistical outliers (\eg linear regression).

There are other ways how samples can ``stand out'' and hence be interesting to look at.
For example, the sample which is closest to the average over all samples can be seen as most representative of the whole set,
or, if there are different output scores, it is sensible to look at a few samples with different scores.
Other sophisticated methods exist to obtain representative and diverse examples for visualizing sample collections.
For image collections, summarization is most commonly done by selecting representative examples.
For example, in \cite{Tschiatschek:2014:LMS:2968826.2968984} the selection of representative images is formulated as optimization problem and mixtures of submodular functions are learned for scoring selections.
In \cite{ZHAO201648} the authors extract SIFT features and use a modification of RANSAC \cite{fischler1981random} plus Affinity Propagation clustering \cite{frey2007clustering} for finding representative images.
If there is accompanying textual or social information for the images, other approaches exist (\eg see \cite{Samani2017, camargo2016multimodal, jaffe2006generating}).

\subsubsection*{Text summarization}
For textual data, visualization and summarization techniques have been extensively surveyed \cite{das2007survey, nenkova2012survey, kucher2015text, gambhir2017recent, allahyari2017text},
and it is commonly distinguished between extractive techniques 
and abstractive techniques. 
Extractive summarization techniques aim at compiling a list of sentences (examples) that summarize the collection.
Abstractive summarization techniques include the extraction of topic words, 
frequency-driven approaches such as tf-idf,
and automatic summarization. 
It is important to note that in our context, we generally not only want to summarize all the given inputs,
but summarize in a way that reveals differences between between inputs associated with different outputs.
Specific works on discriminative text summarization include
\cite{wang2012comparative}, which explains how to select discriminative sentences for summarizing differences between text collections,
and \cite{DBLP:journals/corr/HornAMMS17}, which aims at visualizing differences between text corpora based on discriminative words or by analyzing an SVM that was trained to detect the source of the text.

\subsubsection*{Usage}
Note that ultimately, in interpretation analysis we are not interested in merely visualizing the collection of inputs or outputs,
but to do so in a way that shows relations between input and output values. 
There are three main ways how this can be achieved:
1) If we want to apply dimensionality reduction to the input, the associated values can directly be incorporated into the visualization, \eg by using colors to indicate different associated output values.
2) For applying dimensionality reduction to the output, if we have (short) text data or images as input data it is possible to show the original inputs at the locations of their corresponding output embeddings. 
3) Finally, for example-based approaches and text summarization, input samples can be partitioned based on associated values for separate visualization and successive comparison of results.

Visualization techniques can be very beneficial for an intuitive understanding of perspective,
and can serve as useful starting point for getting ideas about which features to explore
or which types of hypotheses to test with quantitative methods.
On the downside, it is hard to draw any concrete conclusions from visualizations alone.

\section{Comparing multiple perspectives} \label{sec:compare}

Understanding differences between various perspectives has use-cases in a variety of scenarios.
For example, one might be interested in the difference between two given machine learning classifiers, 
understanding how distinct annotators label data differently,
comparing a classifier's perspective to the ground truth human perspective,
or analyzing in which ways data from different domains relates to interpretation-related discrepancies.
Even if one is not directly interested in such a comparative study
and the ultimate interest is only in understanding one given perspective,
it is sensible to compare against a baseline perspective for making results easier to interpret.
For example, it seems that the user in our toy example (Table~\ref{tab:pattern_mining}) slightly prefers explosions in images, but perhaps everyone has such a preference? Maybe what's really special about this user's interpretation is that nudity or humor do not seem to affect her preference in clear way? 

\begin{figure}[h]
\begin{center}
   \includegraphics[width=0.8\linewidth]{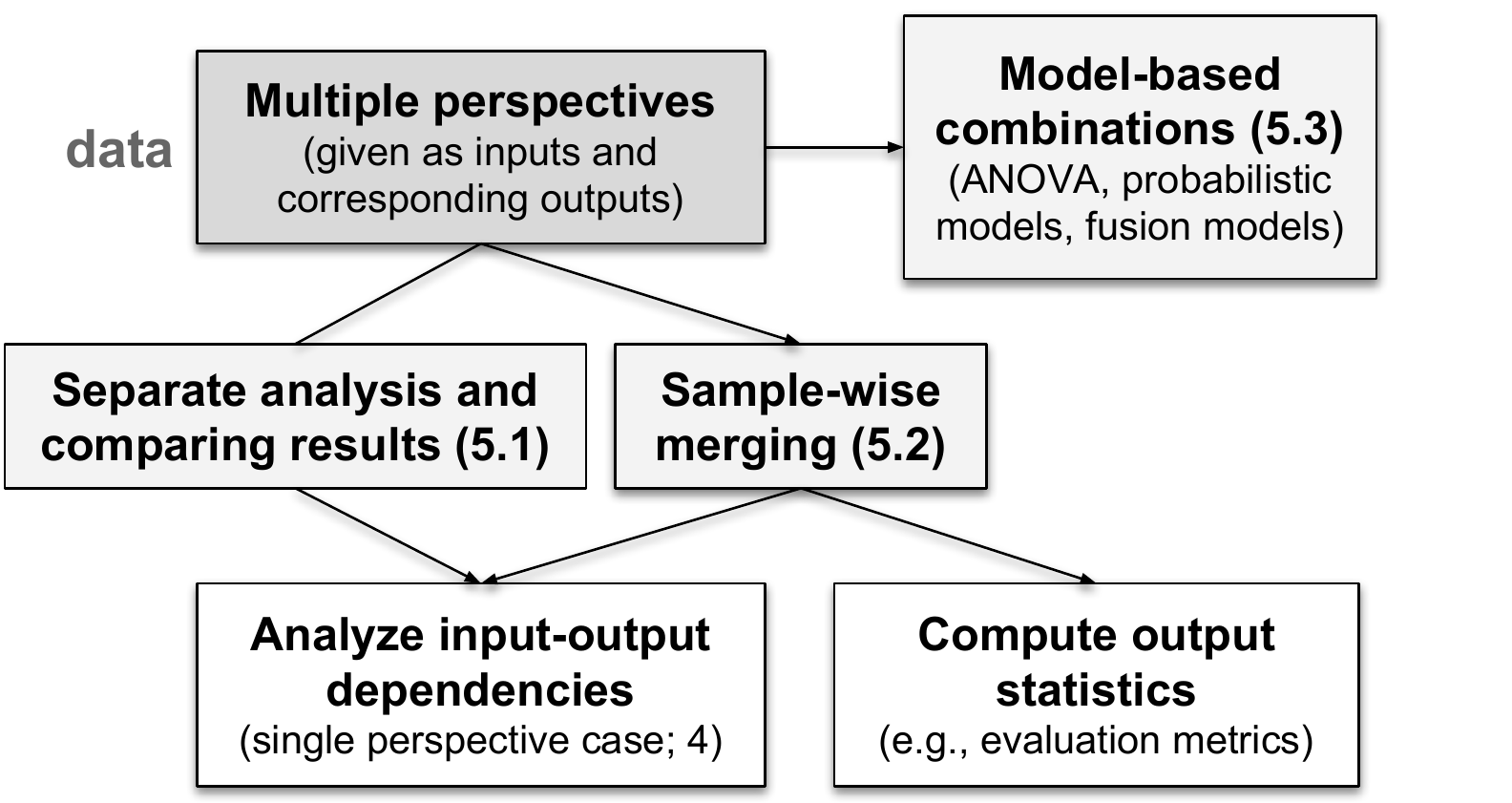}
\end{center}
   \caption{Approaches for comparing multiple ways of interpretation. We can distinguish between three possibilities, out of which two mainly reduce the comparison problem to the analysis of a single perspective.}
\label{fig:comparing}
\end{figure}

So, assume we are given several lists of inputs and their corresponding outputs, each list being associated to one perspective,
and we want to characterize in which ways the underlying interpretation functions are different.
For example, in our image preference scenario, we can imagine to be given similar tables from other users and want to see how their preferences differ.
To this end, individual perspectives can be analyzed separately and then compared, one can merge the perspectives into a single one and then analyze, or combine all perspectives in a single model.
We discuss all of these possibilities for comparison below.
An overview can be found in Figure~\ref{fig:comparing}.

\subsection{Comparing input-output dependencies} \label{sec:compare_dependencies}

Most recent papers that aim at explaining differences of machine learning models
first analyze input-output dependencies by using model-based approaches mentioned above (Section~\ref{sec:model_based}),
and then compare the results, typically by displaying them side by side (see \cite{MONTAVON20181}).
Such an approach of separately analyzing individual perspectives followed by comparison can be seen as direct attempt to answer the question ``How do relations between inputs and outputs differ across the given perspectives?''

Conceptually this offers a simple way to compare,
but can suffer from several issues:
Findings for the different ways of interpretation might be very similar and differences not at all apparent.
For instance, if for user A we find a single rule ``nudity and explosions lead to like in 70\% of cases'' while for user B we obtain ``explosions lead to like in 60\% of cases'', then what exactly is the difference between their ways of interpretation?
Also, if there are many interpretation functions, but only little data for each, analyzing individual perspectives might be unfeasible or not give any significant results.
For example, in the context of recommender systems we might only have 5 items rated per user, which is insufficient for complex statistical analyses or using model-based approaches (on a single user).

Despite these potential shortcomings, there are cases where it makes perfect sense to analyze perspectives separately and then compare.
Most importantly, often there is an interest in understanding individual ways of interpretation as well.
In such cases, individual perspectives would typically be analyzed anyway, so comparing results would only cause little computational overhead and thus provides a reasonable starting point. 
When facing any of the above-mentioned issues, one can still follow up with sample-wise or model-based combinations, which we discuss in the remainder of this section.

\subsection{Sample-wise combinations} \label{sec:sample_wise}
We can phrase the slightly different question ``How do inputs relate to differences in the outputs?''
Let us first assume we have function values from two different interpretation functions $f_{b_1},f_{b_2}$ on the same set of input samples $i_1,\ldots,i_n$.
We can easily define a new perspective $f$ that is described by the same input samples and their associated outputs $d(f_{b_1}(i_1),f_{b_2}(i_1))$, $\ldots$, $d(f_{b_1}(i_n),f_{b_2}(i_n))$,
where $d$ is any real-valued vector function that calculates a difference or distance between two values, \eg $d(y_1,y_2) = | y_1 - y_2 |$.
Thereby, the function $d$ should be chosen depending on the overall goal:
If one is only interested in finding out explanations for when there is disagreement between the two perspectives, one might want to choose a binary indicator of equality,
or the absolute value of the difference between both outputs.
If the goal is to also understand the direction of disagreement, the mere difference without absolute value is more suitable.
For example, if we are given two computer models A and B for sentence-level sarcasm detection, we might ask which features of the sentence are related to any disagreement between A and B (binary case),
but we can also analyze which features make model A but not B vote for sarcasm.

Irrespective of the choice of the merging function $d$, this resulting perspective $f$ can be analyzed as in the single perspective case.
This is a straight-forward way to directly analyze differences between ways of interpretation, and checking statistical significance works in the same way as for a single perspective.
For such a merged perspective, 
output statistics can be computed too, for example in order to evaluate a learned perspective $f_{b_1}$ against a target perspective $f_{b_2}$.
The case of comparing more than two interpretation functions can be handled analogously.

\subsubsection*{Remark -- performance measures}
Many performance measures can be seen as sample-wise combination approaches, 
where the perspective of the classifier is compared to the perspective given by the ground truth labels.
Typically these measures combine the perspectives 
in fairly simple ways. For example, accuracy would use a binary equality indicator as $d$ and average over all outputs of the merged perspective, precision would use the same $d$ but average only over a certain part of the outputs (the ones where the interpretation of the classifier was positive).

\subsection{Model-based combinations} \label{sec:model_combination}
Another possibility is to combine several perspectives in a single model. 
This relates to the question ``How does the bearer influence interpretation?''
Models for model-based combination of perspectives can take various forms, three of which we are going to discuss in this section.

\subsubsection*{ANOVA}
A simple case would be the use of ANOVA, with interpretation output as dependent variable
and both input features and identifier of the interpretation function (or features that group them, such as demographic information) as independent variables.
ANOVA would then tell us whether there is a significant difference among average output values across the perspectives.

\subsubsection*{Probabilistic models}
Even though less common, there are more complex possibilities for combining perspectives in probabilistic models.
Typically, the main goal of such probabilistic models 
is not to analyze ways of interpretation, but to learn how to combine multiple perspectives for a given prediction task.
Still, characteristic information about the involved perspectives can be picked up by such models. 
An example of such a probabilistic model for combining human perspectives is the Dawid-Skene model \cite{dawid1979maximum},
which unites observations from different sources while estimating the observers' errors.
Further examples will be given in the application section (Section~\ref{sec:annotation}).

\subsubsection*{Fusion models}
For AI approaches, ensemble methods are frequently used for increasing predictive performance \cite{rokach2010ensemble}.
These methods often include a scoring mechanism or allow for similar ways of obtaining an estimation of the usefulness of the individual models involved,
which can be seen as discriminative characterization.

\begin{figure}[bt]
\begin{center}
   \includegraphics[width=0.6\linewidth]{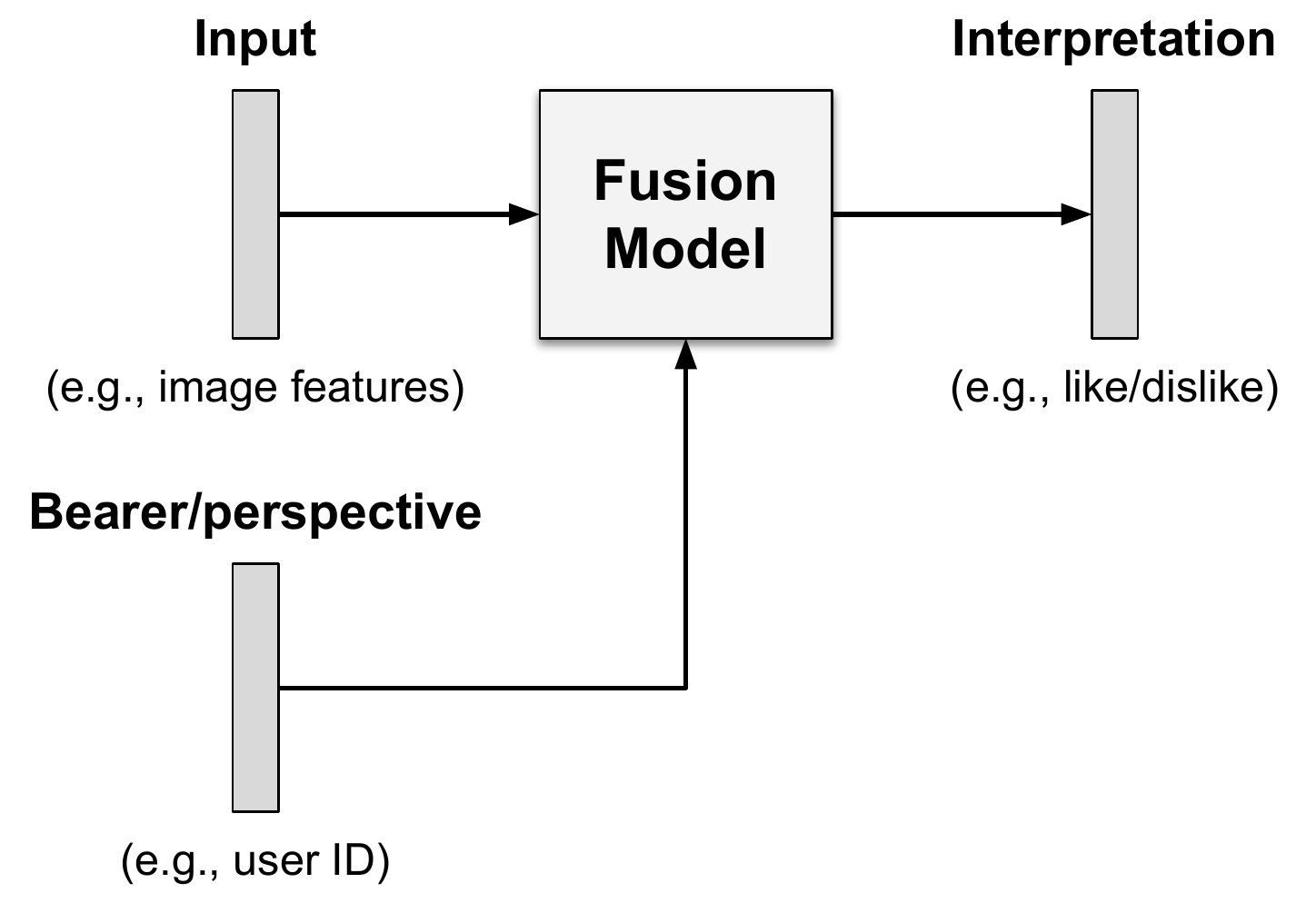}
\end{center}
   \caption{Illustration of end-to-end fusion models for comparing perspectives. Such models are first trained to predict interpretation results from input and information about the bearer. The trained model is then used for analysis.}
\label{fig:end_to_end}
\end{figure}

Another approach to fusion is taken in end-to-end fusion models, where a single model (usually a neural network) is trained to predict the interpretation result given the input and information about the bearer.
This end-to-end approach is illustrated in Figure~\ref{fig:end_to_end}.
This way of combining information is relatively common for prediction but has rarely been used for the purpose of analysis.
Possibilities for analysis include heatmapping and prototype techniques (see Section~\ref{sec:model_based}).
Additionally, end-to-end fusion offers extra opportunities such as learning vector representations for the individual perspectives
which can be used for clustering for example.
However, extra care should be taken when interpreting findings based on such complex models.


\section{Applications} \label{sec:applications}
Tools for interpretation analysis can be utilized in a variety of scenarios.
In the following, we outline some of the cornerstones.

\subsection{Mining subjective information}
Prominent examples of applications that aim at mining subjective information from text data are
sentiment analysis and opinion mining \cite{liu2012survey, ravi2015survey}. 
The main task of sentiment analysis is to decide whether a given text expresses a positive, a negative, or a neutral opinion,
which can for example be useful for evaluating customer reviews.
In its original form, sentiment analysis is about learning a way of interpretation, but does not necessarily involve any claims about characteristics of the same.
However, it is very common to not simply detect overall sentiment, but to do so based on aspects.
The resulting detection pipeline then has aspect information as extra component, and tries to explain the overall sentiment in terms of mentioned aspects and the orientation expressed towards these. 
For understanding persisting differences in interpretation, contrastive opinion mining has been proposed by Fang et al. \cite{Fang:2012:MCO:2124295.2124306} and, later, perspective detection by Vilares and He \cite{vilares2017detecting}.
The Latent Argument Model in \cite{vilares2017detecting} is a rather complex case of discriminative text summarization based on topic modeling,
and is paired in the paper with selection of characteristic sentences.
%
Note that sentiment analysis was extended to the visual modality as well.
Somewhat similar to aspect-based sentiment detection, Borth et al. proposed a visual sentiment ontology \cite{VSO} consisting of adjective-noun combinations (\eg ``scary dog'', ``cute baby'')
that are visually detectable and can be used for explaining the overall sentiment of an image.

Quite a different approach is taken in \cite{li2015identifying}, which analyzes how hotel preferences change over time by applying emerging pattern mining on hotel features mentioned in online reviews.

\subsection{Model analysis}
Several papers have explored the possibility to use decision trees for explaining more complex machine learning models, including neural networks \cite{andrews1995survey, KRISHNAN19991999, boz2002extracting} and tree ensembles \cite{chipman1998making, tan2016tree}.
%
Furthermore, much recent work was done on analyzing deep learning models and explaining decisions based on heatmapping (\eg \cite{zintgraf2017, montavon2017explaining, kindermans2017patternnet, kindermans2017learning}).
These are all direct cases of model-based interpretation analysis (usually not operating under the same black-box assumption though).
Visualization techniques have been used as well for examining learned representations of neural networks.
For example, \cite{cho2014learning} use t-SNE on phrase embeddings (which can be seen as output of the model's interpretation function)
to analyze how semantically meaningful the learned embeddings are.

Note that computation of many performance metrics can be seen as special case of interpretation analysis,
where the output of a classifier is compared to a ground truth human interpretation by merging both perspectives in a sample-wise manner and then aggregating over the outputs of this combined perspective.

\subsection{Annotation} \label{sec:annotation}
Computer vision in particular depends on big amounts of manually labelled data for training models, which is often achieved via crowdsourcing \cite{kovashka2016crowdsourcing}.
In crowdsourcing, it is common to collect several annotations for each item,
and many probabilistic models for merging annotator votes have been proposed (\eg \cite{raykar2010learning, rodrigues2013learning, yan2010modeling, branson2017lean, Horn_2018_CVPR}).
Often, these simultaneously estimate annotator reliability, but only a few approaches consider item difficulty
and thereby relate disagreements to the input.
Notable exceptions are \cite{branson2017lean} and its extension \cite{Horn_2018_CVPR}, which describe such a probabilistic framework and apply their framework to merge fine-grained bird image annotations.
%
Less work has been done on investigating where annotator disagreements come from. 
One of the few examples in this direction is \cite{patton2019annotating}, which analyzes correlations between textual and visual item features and annotator disagreement in case of labeling multimodal tweets as \textit{aggression}, \textit{loss} and \textit{substance use}.
For crowdsourcing, Eickhoff \cite{eickhoff2018cognitive} outlines several quality issues and performs dedicated experiments for analyzing cognitive biases of annotators.
The paper also shows how such biases can propagate into model evaluation and hence have detrimental consequences,
which gives reason for further investigation into a more fine-grained interpretation analysis for annotation.

\subsection{Data understanding and expertise}
In many scientific undertakings the goal is to understand the relation between two quantities based on some given data.
Interpretation analysis tools have been applied to make sense of various kinds of scientific data.
%
Early examples include the application of CCA to describe the relation between wheat to flour characteristics \cite{waugh1942regressions} or to analyze how housing quality interacts with mental issues \cite{hopkins1969statistical}. 
As another example, association rule mining has been used for making sense of gene expression data \cite{creighton2003mining} and medical data \cite{ordonez2006constraining}.
Emerging pattern mining for finding differences between toxical vs non-toxical chemicals \cite{sherhod2014emerging}.
Visual pattern mining for histology image collections is done in \cite{cruz2011visual} for identifying local features that can be used to discriminate between tissue types. 
The same paper also estimates posterior probabilities for relating local features to individual tissue types for interpretation.
Numerous attempts at data explanation have also been made by fitting various models on the given data and then analyzing the trained models for insights.
In \cite{schutt2017quantum}, a deep tensor neural network model with heatmapping was applied to examine the link from molecular structure to electronic properties. 
And \cite{sturm2016interpretable} reported LRP-based explanations for classifying EEG data with a neural network to be highly plausible.
Essentially, such cases can be seen as figuring out some ``natural'' way of interpretation that is intrinsic to the given data. 
In the special case when the output quantity is given in the form of labels from human experts,
analyzing the data amounts to explaining their expert view,
or in other words, to characterize an expert's way of interpretation.
Note, however, that for data understanding our black-box assumption (see Section~\ref{sec:assumptions}) is generally satisfied,
so care has be taken when interpreting the trained model.

\subsection{Understanding human interpretation}
Mechanisms and properties of human interpretation are of fundamental interest in several fields,
including cognitive science, neuroscience, phenomenology, linguistics, psychology and psychiatry.
Traditionally, these fields often conduct designated controlled experiments for data collection,
or use qualitative analysis when relying on given observational data.
%
Still, there are some approaches that are more in between the fields mentioned above and computer science.
These include recent works on computational psychiatry \cite{MONTAGUE201272, stephan2014computational, adams2016computational} which turn hypothesis about human functioning into simple computational models
that can be evaluated on experimental or observational data.
For example, \cite{adams2016computational} explains how to use a hierarchical generative model for exploring potential relations between over-attention to low-level stimuli and schizophrenia.
Another model-based approach is taken in \cite{kawai2018computational} for studying language acquisition by feeding language data into a model based on hidden Markov model. Their trained model is then evaluated by comparing the model's word generalization abilities against the ones of children \cite{imai2008novel}, and can be useful for generating predictions about language development.

\section{Conclusion} \label{sec:conclusion}

In this paper, we proposed a theoretical framework in which we formally defined interpretation, perspective and the task of interpretation analysis.
In our framework, interpretation analysis can be understood as characterizing functions and describes relations between inputs and corresponding outputs.
We showed how analyzing a single way of interpretation can be approached under the use of statistical methods, pattern mining techniques, model-based approaches and visualization techniques.
We discussed how comparing several ways of interpretation can often be reduced to the single perspective case, and alternatively be handled by uniting perspectives in a designated model for analysis.
Finally, we have seen applications from several areas, including opinion mining, annotation and analysis of machine learning models,
which can be connected by their relations to interpretation analysis.

During our survey of approaches, we identified several points that we think deserve more attention in the future.
In particular, proper evaluation of interpretation analysis methods is still largely an open issue.
This holds true especially for more complex model-based approaches under our black-box assumption
(generally satisfied when using them for data understanding)
and visualization techniques.
Further, there are many qualitative methods that are relevant to interpretation analysis which we hope can further inspire computational methods in the future. 
Similarly, though we have already drawn many connections between literature from the fields of behavioural sciences, psychology and computer science in this paper, we hope to see more work in the fruitful intersection of these fields in the future.
Last but not least, we see an ever increasing need for ethical discussions:
Many application areas of interpretation analysis ethically concern user privacy.
Similar techniques to the ones described have in the recent past already been used for ethically very questionable goals under the term microtargeting (\eg to influence the outcome of elections \cite{zuiderveen2018online}).
Our hope is that the scientific community will in the future focus on using the same techniques for ethically less questionable goals, for example to increase transparency and explainability of AI systems and maybe even to help us become aware of our own detrimental biases.

\section*{Acknowledgments}
This work was supported by the BMBF project DeFuseNN (grant number 01IW17002) and the NVIDIA AI Lab (NVAIL) program. Furthermore, the first author received financial support from the Center
for Cognitive Science, Kaiserslautern, Germany.

\appendix



\bibliographystyle{elsarticle-num} 
\bibliography{main.bib}





\end{document}